\documentclass{article}

\PassOptionsToPackage{numbers, compress}{natbib}

\usepackage[preprint]{neurips_2021}

\usepackage[utf8]{inputenc} 
\usepackage[T1]{fontenc}    
\usepackage{hyperref}       
\usepackage{url}            
\usepackage{booktabs}       
\usepackage{amsfonts}       
\usepackage{nicefrac}       
\usepackage{microtype}      
\usepackage{xcolor}         

\usepackage{graphicx}
\usepackage{subfigure}
\usepackage{booktabs} 
\usepackage{hyperref}
\usepackage{amsmath}
\usepackage{comment}
\usepackage{pdflscape}
\usepackage{algorithm}
\usepackage{algorithmic}
\usepackage{bibspacing}
\usepackage{bm}
\usepackage{soul}
\usepackage{multirow}
\usepackage{multicol}
\usepackage{subfigure}
\usepackage{pdflscape}
\usepackage{color,soul}
\usepackage[normalem]{ulem}
\usepackage{wrapfig}

\usepackage{listings}
\usepackage{bbm}

\setcitestyle{square,comma}

\newcommand\Tstrut{\rule{0pt}{2.2ex}}         

\title{Gaussian Process Meta Few-shot Classifier Learning via Linear Discriminant Laplace Approximation
}

\author{%
  Minyoung Kim$^1$
  \\
  $^1$Samsung AI Center \\
  Cambridge, UK \\
  \texttt{mikim21@gmail.com} \\
  \And
  Timothy Hospedales$^{1,2}$ \\
  $^2$University of Edinburgh\\
  Edinburgh, United Kingdom\\
  \texttt{t.hospedales@ed.ac.uk} \\
}

\begin{document}

\maketitle

\begin{abstract}
The meta learning few-shot classification is an emerging problem in machine learning that received enormous attention recently, where the goal is to learn a model that can quickly adapt to a new task with only a few labeled data. 
We consider the Bayesian Gaussian process (GP) approach, in which we meta-learn the GP prior, 
and the adaptation to a new task is carried out by the GP predictive model from the posterior inference. 
We adopt the Laplace posterior approximation, but to circumvent the iterative gradient steps for finding the MAP solution, we introduce a novel linear discriminant analysis (LDA) plugin as a surrogate for the MAP solution. 
In essence, the MAP solution is approximated by the 
LDA estimate, but to take the GP prior into account, we adopt the prior-norm adjustment to estimate LDA's shared variance parameters, which ensures that the adjusted estimate is consistent with the GP prior. This enables closed-form differentiable 
GP posteriors and predictive distributions, thus allowing fast meta training. 
We demonstrate 
considerable improvement over the previous approaches. 
\end{abstract}

\section{Introduction}\label{sec:intro}

{\em Few-shot classification}~\citep{oneshot,survey_few_shot,survey_meta,survey_deep_meta} is the task of predicting class labels of data instances that have novel unseen class semantics, potentially from a novel domain, where the learner is given only a few labeled data from the domain. It receives significant attention recently in machine learning, not only due to the practical reason that annotating a large amount of data for training deep models is prohibitively expensive, but also the constant endeavor in AI to build human-like intelligence where the human is extremely good at recognizing new categories from a few examples. 

In order to build a model that can generalize well to a novel task with only a few samples, 
{\em meta learning}~\citep{meta_schmid,meta_bengio} forms a training stage that is similar to the test scenario. More specifically, during the training stage, the learner sees many tasks (or episodes) where each task consists of the {\em support} and {\em query} sets: the learner adapts the model to the current task using a few labeled data in the support set, and the performance of the adapted model is measured on the query set, which is used as a learning signal to update the learner. This is in nature a learning-to-learn paradigm~\citep{learn2learn,protonet,matchingnet,maml}, and it often leads to more promising results in certain scenarios than simple supervised feature (transfer) learning. Although recently there were strong baselines introduced for the latter with some feature transformations~\citep{simpleshot,baseline}, in this paper we focus on the meta learning paradigm. 

As meta few-shot learning essentially aims to generalize well from only a few observations about a new task domain, it is important to learn prior information that is shared across different tasks. In this sense, the Bayesian approach~\citep{prob_semantic,grow_mind} is attractive in that we can express the prior belief effectively, and easily adapt our belief to a new task based on the given evidence, in a principled manner. In Bayesian meta learning~\citep{llama,bmaml,versa,meta_mix,metafun}, the adaptation to a new task corresponds to posterior predictive distribution inference, and meta learning amounts to learning a good prior distribution from many training episodes. 

To enable efficient Bayesian meta learning, the posterior predictive inference needs to be fast and succinct (e.g., closed form). To this end, we consider the Gaussian process (GP) model with the linear deep kernel~\citep{dkl16} that allows parametric treatment of GP via the weight-space view~\citep{gpml_book}. Although there was similar attempt recently~\citep{gpdkt}, they resort to regression-based likelihood model for the classification problem to derive  closed-form inference, and such an ad hoc strategy can potentially lead to performance degradation. 
Instead, we propose a novel Laplace approximation for the GP posterior with a linear discriminant plugin, which avoids iterative gradient steps to find the maximum-a-posterior (MAP) adaptation solution, and allows a closed-form predictive distribution that can be used in stochastic gradient 
meta training efficiently. Hence, it is computationally more attractive than gradient-based adaptation  approaches~\citep{maml,llama,meta_mix} by construction, and more amenable to train than  neural net approximations of the predictive distribution (i.e., amortization) such as~\citep{versa}. 

We show the improved performance of our GP approach over the regression-based previous work~\citep{gpdkt} and other state-of-the-arts on several benchmark datasets in both within- and cross-domain few-shot learning problems.

\section{Problem Setup and Background}\label{sec:background}

We provide the formal training/test setup for the meta few-shot classification problem (Sec.~\ref{sec:setup}). 
We then briefly review the recent GP-based few-shot learning algorithm (GPDKT)~\citep{gpdkt} in Sec.~\ref{sec:gpdkt} due to its close relation to our proposed approach.

\subsection{Meta few-shot learning framework}\label{sec:setup}

The ($C$-way, $k$-shot) episodic meta few-shot classification problem can be formally defined as follows:
\begin{itemize}
\item Training stage (
repeated for $T$ times/episodes):
  \begin{enumerate}
  \item Sample training data $(S,Q)$ for this episode: support set $S = \{(x,y)\}$ and query set $Q = \{(x,y)\}$, where $S$ consists of $C\cdot k$ samples ($k$ samples for each of the $C$ classes), and $Q$ contains $C \cdot k_q$ samples ($k_q$ samples per class). 
  We denote by $y\in\{1,\dots,C\}$ the class labels in $(S,Q)$, however, the semantic meaning of the classes is different from episode to episode. 
  \item With 
  $(S,Q)$, we train a meta learner $\mathcal{F}(S) \to h$ where the output of $\mathcal{F}$ is a $C$-way classifier, 
  $h:\mathcal{X}\to\{1,\dots,C\}$. The training objective 
  is typically defined on the query set, e.g., the prediction error of $h$ on $Q$.   
  \end{enumerate}
\item Test stage: 
  \begin{enumerate}
  \item The $k/k_q$-shot test data $(S^*,Q^*)$ are sampled, but the query set $Q^*$ is not revealed. For the $k$-shot support set $S^* = \{(x,y)\}$, we apply our learned $\mathcal{F}$ to $S^*$ to obtain the classifier $h^* = \mathcal{F}(S^*)$. Again, the semantic meaning of the test class labels are different from those in the training stage. 
  The performance of $h^*$ is measured on the test query set $Q^*$.  
  \end{enumerate}
\end{itemize}

For instance, in the popular ProtoNet~\citep{protonet}, the meta learner learns the parameters $\theta$ of the feature extractor $\phi_\theta(x)$ (e.g., convolutional networks), and the meta learner's output $h = \mathcal{F}(S)$ is the nearest centroid classifer where the centroids are the class-wise means in $S$ in the feature space.  Note that $h(x)$ admits a closed form (softmax), and the meta training updates $\theta$ by stochastic gradient descent with the loss,  $\mathbb{E}_{(x,y)\sim Q} [\textrm{CrossEnt}(y,h(x))]$. 
Another example is the GP meta learning framework that essentially considers $h = \mathcal{F}(S)$ as a GP posterior predictive model, that is, 
\begin{align}
p(y|x,S) = \int p(y|f(x)) \ p(f|S) \ df 
\label{eq:gp_predictive}
\end{align}
where $f$ is a GP function, $h(x)$ is defined as a probabilistic classifier $p(y|x,S)$, and $p(f|S) \propto p(f) \cdot \prod_{(x,y)\in S} p(y|f(x))$. 
Meta training of $\mathcal{F}$ amounts to learning the GP prior distribution $p(f)$ (i.e., GP mean/covariance functions). The recent GPDKT~\citep{gpdkt} is one incarnation of this GP framework. 

\subsection{Brief review of GPDKT
}\label{sec:gpdkt}

GPDKT assumes the GP {\em regression} model (its usage to classification will be described shortly), 
\begin{align}
&f(\cdot) \sim \mathcal{GP}(0, k_\theta(\cdot,\cdot)), \\
&y = f(x) + \epsilon, \ \ \epsilon \sim \mathcal{N}(0,\sigma^2),
\end{align}
where the GP covariance function $k_\theta$ is defined as the deep kernel~\citep{dkl16}:
\begin{align}
k_\theta(x,x') = \tilde{k}(\phi_\theta(x), \phi_\theta(x')),
\label{eq:dkl}
\end{align}
where $\phi_\theta(x)$ is the feature extractor (comparable to that in ProtoNet) 
and $\tilde{k}(\cdot,\cdot)$ is a conventional kernel function (e.g., Gaussian RBF, linear, or cosine similarity). We abuse the notation to denote by $\theta$ all the parameters of the deep kernel, including those from the outer kernel $\tilde{k}$. 
They pose the meta training 
as the marginal likelihood maximization on both support and query sets: 
\begin{align}
\max_\theta \ 
\int p(f) \cdot
\prod_{(x,y)\in S\cup Q} 
p(y|f(x)) \ df. 
\label{eq:gpdkt_marginal}
\end{align}
Due to the regression model, the marginal likelihood admits a closed form, 
and one can easily optimize (\ref{eq:gpdkt_marginal}) by stochastic gradient ascent. 

To extend the GP model to the classification problem, instead of adopting a softmax-type likelihood $p(y|f(x))$, they rather stick to the GP regression model. This is mainly for the closed-form posterior and marginal data likelihood. In the binary classification problem, they assign {\em real-valued} $y=\pm 1.0$ as target response values for positive/negative classes, respectively, during training. At the test time, they threshold the real-valued outputs to get the discrete class labels. For the multi-class $C$-way problem with $C>2$, they turn it into $C$ binary classification problems by one-vs-rest conversion. Then during training, they maximize the sum of the marginal log-likelihood scores over the $C$ binary problems, while at test time the one with the largest predictive mean $\mathbb{E}[y|x,S]$ over the $C$ problems is taken as the predicted class. Although this workaround allows fast adaptation and training with the closed-form solutions from GP regression, the ad hoc treatment of the discrete class labels may degrade the prediction accuracy.

\section{Our Approach
}\label{sec:main}

In this section we describe our Laplace approximation GP posterior formulation for the task adaptation, where we introduce the novel {\em linear discriminant plug-in} to circumvent the iterative optimization for the MAP solution and enable the closed-form formulas. Our formalism admits the softmax classification likelihood model, more sensible than the regression-based treatment of the classification problem.



We adopt the weight-space view of the Gaussian process model~\citep{gpml_book} using the linear-type deep kernel, 
and consider the softmax likelihood model with $C$ functions $F(x) = \{f_j(x)\}_{j=1}^C$:
\begin{align}
& p(y|F(x)) = \frac{e^{f_y(x)}}{\sum_{j=1}^C e^{f_j(x)}}, \ f_j(x) = w_j^\top \phi(x) + b_j, \label{eq:gpldla_lik} \\
& w_j \sim \mathcal{N}(0,\beta^2 I), \ \ b_j \sim \mathcal{N}(0, \beta_b^2) \ \ \textrm{for} \ \  j=1\dots C. \label{eq:gpldla_prior}
\end{align}
We let $W=[w_1,\dots,w_C]$ and $B=[b_1,\dots,b_C]$ be the weight-space random variables for the GP functions. Note that in (\ref{eq:gpldla_prior}) the prior (scalar) parameters $\beta, \beta_b$ are shared over all $C$ functions, which is reasonable considering that the semantic meaning of classes changes from episode to episode. 
And it is easy to see that the i.i.d.~priors on $(w_j,b_j)$ makes $\{f_j(\cdot)\}_{j=1}^C$ i.i.d.~GPs with a zero mean and the covariance function,
\begin{align}
\textrm{Cov}(f_j(x),f_j(x')) = \beta^2 \phi(x)^\top \phi(x') + \beta_b^2.
\label{eq:gp_cov}
\end{align}
%
This can be interpreted as adopting a linear outer kernel $\tilde{k}(z,z') = z^\top z'$ in the deep kernel (\ref{eq:dkl}) with some scaling and biasing. Although our formulation excludes more complex nonlinear outer kernels (e.g., RBF or polynomial), it was shown that the linear or cosine-similarity outer kernel empirically performed the best among other choices~\citep{gpdkt}. Note that the latter cosine-similarity kernel is obtained by unit-norm feature transformation ($\phi(x) \to \frac{\phi(x)}{||\phi(x)||}$). 

Given the support set $S=\{(x,y)\}$, the GP  posterior distribution of $f_j(x)$ at some arbitrary input $x$ becomes $p(f_j(x)|S) = p(w_j^\top\phi(x)+b_j|S)$, and this is determined by the posterior 
$p(W,B|S)$, where (up to constant)
\begin{align}
\log p(W,B|S) 
\ = \ - \sum_{j=1}^C \bigg( \frac{||w_j||^2}{2\beta^2} + \frac{b_j^2}{2\beta_b^2} \bigg) \ +   \sum_{(x,y)\in S} \bigg( w_y^\top\phi(x) + b_y - \log \sum_{j=1}^C e^{w_j^\top\phi(x)+b_j} \bigg). \label{eq:post_wb_given_S} 
\end{align}
The posterior 
$p(W,B|S)$ 
is used to build the task($S$)-adapted classifier $p(y|x,S)$, the GP predictive distribution  derived from (\ref{eq:gp_predictive}). 
And the meta training in our model amounts to optimizing the classification (cross-entropy) loss of the adapted classifier on the query set with respect to the GP prior parameters 
(i.e., $\beta$, $\beta_b$, and the parameters $\theta$ of the feature extractor network $\phi$), 
averaged over all training episodes. That is, our meta training loss/optimization can be written as:
\begin{align}
\min_{\theta,\beta,\beta_b} \mathbb{E}_{(S,Q)} \Bigg[ -\sum_{(x,y)\in Q} \log p(y|x,S) \Bigg], 
\label{eq:meta_train}
\end{align}
where $p(y|x,S) = \iint p(W,B|S) p(y|x,W,B) dW dB$,
and the expectation is taken over $(S,Q)$ samples from training episodes/tasks. 
Considering the dependency of the loss on these prior parameters as per (\ref{eq:meta_train}), it is crucial to have a succinct (e.g., closed-form) expression for 
$p(W,B|S)$, 
as well as the predictive distribution $p(y|x,S)$. 
However, since 
$p(W,B|S)$ 
does not admit a closed form due to the non-closed-form normalizer (i.e., the log-sum-exp of (\ref{eq:post_wb_given_S}) over $\{w_j,b_j\}_j$), 
we adopt the Laplace approximation that essentially approximates (\ref{eq:post_wb_given_S}) by the second-order Taylor at around the MAP estimate $\{w^*_j,b^*_j\}_j$, i.e., the maximum of (\ref{eq:post_wb_given_S}).

\subsection{Laplace approximation via LDA plugin with prior-norm adjustment}

Specifically we follow the diagonal covariance Laplace approximation with diagonalized Hessian of (\ref{eq:post_wb_given_S}), which leads to the factorized posterior 
$p(W,B|S) = \prod_{j=1}^C p(w_j,b_j|S)$.
The approximate posterior can be derived as $p(w_j,b_j|S) \approx \mathcal{N}(w_j; w^*_j, V^*_j) \  \mathcal{N}(b_j; b^*_j, v^*_j)$, with
\begin{align}
&
V^*_j = \textrm{Diag}\bigg( \frac{1}{\beta^2} + \sum_{(x,y)\in S} 
a^*(x,y,j) \phi(x)^2
\bigg)^{-1} \label{eq:laplace_2} \\
&
v^*_j = \bigg( \frac{1}{\beta_b^2} + \sum_{(x,y)\in S} a^*(x,y,j) 
\bigg)^{-1}
\label{eq:laplace_3}
\end{align}
where $a^*(x,y,j) = p(y=j|F^*(x)) - p(y=j|F^*(x))^2$, 
$F^*(x) = \{f^*_j(x)\}_j$ with $f^*_j(x) = {w^*_j}^\top \phi(x) + b^*_j$, and all operations are element-wise. 

However, obtaining the MAP estimate $\{w^*_j,b^*_j\}_j$, i.e., the maximum of (\ref{eq:post_wb_given_S}), although the objective is concave, usually requires several steps of gradient ascent, which can hinder efficient meta training. Recall that our meta training amounts to minimizing the loss of the task-adapted classifier $p(y|x,S)$ on a query set with respect to the feature extractor $\phi_\theta(\cdot)$ and the GP prior parameters $\beta,\beta_b$, and we prefer to have succinct (closed-form-like) expression for $p(y|x,S)$ in terms of $\theta,\beta,\beta_b$. The iterative dependency of $p(y|x,S)$ on $\phi,\beta,\beta_b$, resulting in a similar strategy as MAML~\citep{maml}, would make the meta training computationally expensive. 
To this end, we propose a novel linear discriminant analysis (LDA) plugin technique as a surrogate of the MAP estimate. 

\textbf{LDA-Plugin.} 
We preform the LDA on the support set $S$, which is equivalent to fit a mixture of Gaussians  with equi-covariances by maximum likelihood~\citep{bishop:2006:PRML}. More specifically, we consider the Gaussian mixture model (with some abuse of notation, e.g., $p(x)$ instead of $p(\phi(x))$),  
\begin{align}
p(x,y) = p(y) p(x|y) = \pi_y \mathcal{N}(\phi(x); \mu_y, \sigma^2 I), 
\end{align}
where we assume the shared spherical covariance matrix across different classes. The maximum likelihood (ML) estimate on $S$ can be derived as:
\begin{align}
\pi_j^* = \frac{n_j}{n}, \ 
\mu_j^* = \sum_{x \in S_j} \frac{\phi(x)}{n_j}, \ 
{\sigma^2}^* = \sum_{j=1}^C \sum_{x \in S_j} \frac{||\phi(x)-\mu_j^*||^2}{n d},
\label{eq:lda_estim}
\end{align}
where $S_j = \{ (x,y)\in S: y=j \}$, $n_j = |S_j|$, $n = |S|$, and $d = \textrm{dim}(\phi(x))$. 
Then our idea is to use this ML-estimated Gaussian mixture to induce the class predictive model $p(y|x) = p(x,y)/p(x)$, and match it with our GP likelihood $p(y|F(x))$ in (\ref{eq:gpldla_lik}) to obtain $\{w_j,b_j\}_j$, which serves as a surrogate of the MAP estimate $\{w^*_j,b^*_j\}_j$. More specifically, the class predictive from the Gaussian mixture is:
\begin{align}
&p(y|x) = \frac{\pi_y \mathcal{N}(\phi(x); \mu_y, \sigma^2 I)} {\sum_j \pi_j \mathcal{N}(\phi(x); \mu_j, \sigma^2 I)} = \frac{\exp\big( (\mu_y/\sigma^2)^\top \phi(x) + \log\pi_y - ||\mu_y||^2/(2\sigma^2) \big)} {\sum_j \exp\big( (\mu_j/\sigma^2)^\top \phi(x) + \log\pi_j - ||\mu_j||^2/(2\sigma^2) \big)}. 
\label{eq:lda_class_pred} 
\end{align}
We match it with the GP likelihood model $p(y|F(x))$ from (\ref{eq:gpldla_lik}), that is,
\begin{align}
p(y|F(x)) = \frac{\exp\big( w_y^\top \phi(x) + b_y \big)} {\sum_j \exp\big( w_j^\top \phi(x) + b_j \big)},
\label{eq:gpldla_class_pred}
\end{align}
which establishes the following correspondence: 
\begin{align}
w_j = \frac{\mu_j}{\sigma^2}, \ \ \ \ 
b_j = \log \pi_j - \frac{||\mu_j||^2}{2\sigma^2} + \alpha,
\label{eq:lda_co}
\end{align}
where $\alpha$ is a constant (to be estimated). We aim to plug the LDA estimates 
(\ref{eq:lda_estim})
in (\ref{eq:lda_co}), to obtain the MAP surrogate. 
However, there are two issues in this strategy: First, the ML estimate ${\sigma^2}^*$ can raise a numerical issue in the few-shot learning since the number of samples is too small\footnote{In the one-shot case ($n_j=1$), e.g., 
degenerate 
${\sigma^2}^*=0$.
}, although $\pi^*$ and $\mu^*$ incur no such issue. Secondly, it is only the ML estimate with data $S$, and we have not taken into account the prior on $\{w_j,b_j\}_j$. To address both issues simultaneously, we propose a prior-norm adjustment strategy, which also leads to a sensible estimate for $\sigma^2$. 


\textbf{Prior-norm adjustment.} 
We will find $\sigma^2$ that makes the surrogate $w_j$ in (\ref{eq:lda_co}) become consistent with our prior $p(w_j)=\mathcal{N}(0,\beta^2 I)$. Since $w_j$ sampled from the prior can be written as $w_j = [\beta \epsilon_{j1},\dots,\beta \epsilon_{jd}]^\top$ with $\epsilon_{j1},\dots,\epsilon_{jd} \stackrel{iid}{\sim} \mathcal{N}(0,1)$, we have:
\begin{align}
||w_j||^2 
= \beta^2 \sum_{l=1}^d \epsilon_{jl}^2 = \beta^2 d \cdot \frac{1}{d} \sum_{l=1}^d \epsilon_{jl}^2
\approx \beta^2 d \cdot \mathbb{E}[\epsilon_{jl}^2] = \beta^2 d,
\label{eq:w_norm}
\end{align}
where the approximation to the expectation 
gets more accurate as $d$ increases due to the law of large numbers. The equation (\ref{eq:w_norm}) implies that any $w_j$ that conforms to the prior has the norm approximately equal to $\beta \sqrt{d}$. Hence we enforce this to the surrogate $w_j$ in (\ref{eq:lda_co}) to determine $\sigma^2$. To consider all $j=1\cdots C$, we establish a simple mean-square equation, $(1/C)\sum_{j=1}^C ||\mu_j^*/\sigma^2||^2 = \beta^2 d$,
and the solution leads to the prior-norm adjusted MAP surrogate (denoted by $w_j^*)$ 
as follows:
\begin{align}
{\sigma^2}^* = 
\frac{1}{\beta \sqrt{d}} \sqrt{\frac{1}{C}\sum_{j=1}^C ||\mu_j^*||^2},
\ \ 
\ \ 
w_j^* = \frac{\mu_j^*}{{\sigma^2}^*}.
\label{eq:w_optim}
\end{align}

\textbf{Determining $\alpha$.} 
We adjust $b_j$ to take into account its prior, and from (\ref{eq:lda_co}) this amounts to finding $\alpha$ properly. We directly optimize the log-posterior  (\ref{eq:post_wb_given_S}) with respect to $\alpha$. Denoting $\hat{b}_j = \log \pi_j^* - ||\mu_j^*||^2/(2{\sigma^2}^*)$ 
(i.e.,  $b_j=\hat{b}_j + \alpha$), we solve 
$\frac{\partial \log p(\{b_j\}_j|S)}{\partial \alpha} = -\sum_{j=1}^C \frac{\hat{b}_j+\alpha}{\beta_b^2} = 0$, and 
have a MAP surrogate (denoted by $b_j^*$) as:
\begin{align}
\alpha^* = -\frac{1}{C} \sum_{j=1}^C \hat{b}_j, \ \ \ \ 
b_j^* = \hat{b}_j + \alpha^*.
\label{eq:b_optim}
\end{align}

\textbf{Summary. 
} 
We have derived the Laplace approximated posterior $p(w_j,b_j|S)$ in (\ref{eq:laplace_2}--\ref{eq:laplace_3}) with the MAP surrogate $(w_j^*,b_j^*)$ from (\ref{eq:w_optim}) and (\ref{eq:b_optim}). From this GP posterior, we derive the predictive distribution $p(y|x,S)$ that is used in our meta training~(\ref{eq:meta_train}) as well as meta test. We adopt the Monte Carlo estimate with $M$ (reparametrized) samples from the posterior: 
\begin{align}
&p(y|x,S) \approx \frac{1}{M} \sum_{m=1}^M p(y|x,W^{(m)},B^{(m)}), \ \ 
\label{eq:predictive} \\ 
& \ \ \ \ \ \ \ \ \textrm{where} \ \ w_j^{(m)} = w_j^* + {V_j^*}^{\frac{1}{2}}\epsilon_j^{(m)}, \ 
b_j^{(m)} = b_j^* + \sqrt{v_j^*}\gamma_j^{(m)}
\nonumber
\end{align}
where $\epsilon_j^{(m)}$ and $\gamma_j^{(m)}$ are iid samples from $\mathcal{N}(0,1)$. Note that the approximate $p(y|x,S)$ in (\ref{eq:predictive}) depends on our GP prior parameters $\{\theta,\beta,\beta_b\}$ in a closed form, making the gradient evaluation and stochastic gradient descent training of (\ref{eq:meta_train}) easy and straightforward. 
For the meta testing, we also use the same Monte Carlo estimate. The number of samples $M=10$ usually works well in all our empirical studies. Our approach is dubbed GPLDLA (Gaussian Process Linear Discriminant Laplace Approximation). 
The final meta training/test algorithms are summarized in Alg.~\ref{alg:gpldla}.


\newcommand\inlineeqno{\stepcounter{equation}\ (\theequation)}
\newcommand{\INDSTATE}[1][1]{\STATE\hspace{#1\algorithmicindent}}
\begin{algorithm}[t!]
\caption{GPLDLA meta training and meta test.}
\label{alg:gpldla}
\begin{small}
\begin{algorithmic}
\STATE \texttt{[META TRAINING]}
    \INDSTATE \textbf{Input:} Initial GP prior parameters: $\theta$, 
    $\beta,\beta_b$. 
    \INDSTATE \textbf{Output:} Trained $\theta,\beta,\beta_b$.
    \INDSTATE \textbf{Repeat:}
        \INDSTATE[2] 0. Sample an episode/task.
        \INDSTATE[2] 1. Sample data $(S,Q)$ from the current episode.
        \INDSTATE[2] 2. Estimate 
        $\{\pi_j^*,\mu_j^*\}$ with 
        $S$ using (\ref{eq:lda_estim}).
        \INDSTATE[2] 3. Estimate 
        ${\sigma^2}^*$ and 
        $\{w_j^*,b_j^*\}$ using (\ref{eq:w_optim}--\ref{eq:b_optim}).
        \INDSTATE[2] 4. Update $\theta,\beta,\beta_b$ by SGD with (\ref{eq:meta_train}) using (\ref{eq:predictive}). 
\STATE \texttt{[META TEST]}
    \INDSTATE \textbf{Input:} Trained 
    $\theta,\beta,\beta_b$ and test samples $(S^*,Q^*)$.
    \INDSTATE \textbf{Output:} Predictive distr. $p(y^*|x^*,S^*)$ for $x^*\in Q^*$. 
    \INDSTATE Do 2 $\&$ 3 above with $S^*$; compute $p(y^*|x^*,S^*)$ by (\ref{eq:predictive}).
\end{algorithmic}
\end{small}
\end{algorithm}

\section{Related Work}\label{sec:related}

Few-shot/meta learning~\citep{meta_bengio,meta_schmid} 
has received enormous attention recently with the surge of deep learning, and it now has a large body of literature~\citep{survey_few_shot,survey_meta,survey_deep_meta}. 
The approaches in few-shot learning can broadly fall into two folds: 
feature transfer and the other meta learning. The former uses the entire training data 
to pretrain the feature extractor network, which is then adapted to a new task by finetuning the network or training the output heads with the few-shot test data~\citep{simpleshot,baseline}. On the other hand, the meta learning approaches~\citep{protonet,matchingnet,maml} follow the learning-to-learn paradigm~\citep{learn2learn}, where the meta learner 
is trained by the empirical risk minimization principle. 


In the Bayesian meta learning~\citep{llama,bmaml,versa,meta_mix,metafun}, the prior on the underlying model parameters typically serves as the meta learner, and the adaptation to a new task corresponds to inference of the posterior predictive distribution. In this way the meta learning amounts to learning a good prior distribution from many training episodes. 
For the efficient meta training, the posterior predictive inference, i.e., adaptation procedure, needs to be fast and succinct (e.g., in closed forms). Some previous approaches used neural net approximation of the posterior predictive distribution (i.e., amortization)~\citep{versa,metafun}, while others are based on gradient updates~\citep{maml,llama,meta_mix}. The main focus of meta few-shot learning lies on how to learn the meaningful prior model that can be quickly and accurately adaptable to novel tasks 
with only a limited amount of evidence. 

%
Another recent Bayesian meta learner closely related to ours is MetaQDA~\citep{metaqda}, where they consider a mixture-of-Gaussians (MoG) classifier possibly with non-equal covariances, thus  representing quadratic decision boundaries. With the Normalized-Inverse-Wishart prior on the MoG parameters, the posterior admits a closed-form expression by conjugacy. 
One of the key differences from our approach is that the MetaQDA deals with the joint MoG modeling $p(x,y)$, while we focus on the discriminative $p(y|x)$. It is known that the discriminative model has lower asymptotic error and is more data efficient without requiring marginal input distribution modeling~\citep{ng_jordan}. But this comes at the cost of the non-closed-form posterior, and we had to resort to Laplace approximation with the prior-norm adjusted MAP estimates. Despite superb performance, there are 
several shortcomings of MetaQDA: it involves a large number of Wishart prior parameters to be trained, $O(C d^2)$ for $C$-way classification and $d$-dimensional features. On the other hand, ours has only two extra scalar parameters $\beta,\beta_b$.
Moreover, MetaQDA's performance is rarely known when the backbone feature extractor network $\phi(x)$ is {\em jointly} trained. They rather fix the features and only learn the prior QDA model.  
Its performance is highly reliant on the underlying feature extractor used. 


\begin{table*}
\caption{Average accuracies and standard deviations on the CUB dataset. Best results are boldfaced.}
\vspace{+0.3em}
\centering
\begin{footnotesize}
\centering
\scalebox{0.95}{
\begin{tabular}{lcccc}
\toprule
\multirow{2}{*}{Methods} & 
\multicolumn{2}{c}{Conv-4} &  \multicolumn{2}{c}{ResNet-10}
\\ \cline{2-5}
& 1-shot\Tstrut & 5-shot & 1-shot & 5-shot \\
\hline\hline
Feature Transfer\Tstrut & $46.19 \pm 0.64$ & $68.40 \pm 0.79$ & $63.64 \pm 0.91$ & $81.27 \pm 0.57$ \\
Baseline$++$~\citep{baseline}\Tstrut & $61.75 \pm 0.95$ & $78.51 \pm 0.59$ & $69.55 \pm 0.89$ & $85.17 \pm 0.50$ \\
MatchingNet~\citep{matchingnet}\Tstrut & $60.19 \pm 1.02$ & $75.11 \pm 0.35$ & $71.29 \pm 0.87$ & $83.47 \pm 0.58$ \\
ProtoNet~\citep{protonet}\Tstrut & $52.52 \pm 1.90$ & $75.93 \pm 0.46$ & ${\bf 73.22 \pm 0.92}$ & $85.01 \pm 0.52$ \\
MAML~\citep{maml}\Tstrut & $56.11 \pm 0.69$ & $74.84 \pm 0.62$ & $70.32 \pm 0.99$ & $80.93 \pm 0.71$ \\
RelationNet~\citep{relationnet}\Tstrut & $62.52 \pm 0.34$ & $78.22 \pm 0.07$ & $70.47 \pm 0.99$ & $83.70 \pm 0.55$ \\
SimpleShot~\citep{simpleshot}\Tstrut & $-$ & $-$ & $53.78 \pm 0.21$ & $71.41 \pm 0.17$ \\
GPDKT$^\textrm{CosSim}$~\citep{gpdkt}\Tstrut & $63.37 \pm 0.19$ & $77.73 \pm 0.26$ & $70.81 \pm 0.52$ & $83.26 \pm 0.50$ \\
GPDKT$^\textrm{BNCosSim}$~\citep{gpdkt}\Tstrut & $62.96 \pm 0.62$ & $77.76 \pm 0.62$ & $72.27 \pm 0.30$ & $85.64 \pm 0.29$ \\
\hline
GPLDLA (Ours)\Tstrut & ${\bf 63.40 \pm 0.14}$ & ${\bf 78.86 \pm 0.35}$ & $71.30 \pm 0.16$ & ${\bf 86.38 \pm 0.15}$ \\
\bottomrule
\end{tabular}
}
\end{footnotesize}
\label{tab:cub}
\end{table*}

\section{Experiments}\label{sec:expmt}

In this section we test our GPLDLA on several popular benchmark tasks/datasets in meta few-shot classification. We demonstrate the performance improvement over the state-of-the-arts, especially highlighting more accurate prediction than the previous GP few-shot model, GPDKT~\citep{gpdkt}.

\textbf{Implementation details.} 
For fair comparison with existing approaches, we use the same feature extractor backbone network architectures $\phi_\theta(x)$ (e.g., convolutional networks or ResNets~\citep{resnet}) as competing models such as ProtoNet~\citep{protonet}, Baseline$++$~\citep{baseline},  SimpleShot~\citep{simpleshot}, and GPDKT~\citep{gpdkt}. For all experiments we use normalized features ($\phi(x) \to \frac{\phi(x)}{||\phi(x)||}$), which corresponds to the cosine-similarity outer kernel with the  original feature in our deep kernel GP covariance function (\ref{eq:gp_cov}). As the GP prior parameters $\beta,\beta_b$, the only extra parameters, are constrained to be positive, we represent them as exponential forms and perform gradient descent in the exponent space. The number of Monte Carlo samples is fixed as $M=10$ for all experiments. The details of the optimization hyperparaemters are described in Appendix. 

\textbf{Datasets/tasks and protocols.} 
We consider both within-domain and cross-domain few-shot learning setups: the former takes the training and test episodes/tasks from the same dataset, while the latter takes training tasks from one dataset and test tasks from another. For the within-domain setup, we use the three most popular datasets, the Caltech-UCSD Birds~\citep{cub} (denoted by \textbf{CUB}), the \textbf{{\em mini}ImageNet}~\citep{matchingnet}, and the \textbf{{\em tiered}ImageNet}~\citep{tiered}. The CUB dataset has 11788 images from 200 classes, the miniImageNet has 60,000 images from 100 classes, while the tieredImageNet contains 779,165 images from 608 classes. We follow the standard data split: 100/50/50 classes for training/validation/test data for CUB, 64/16/20 for miniImageNet, and 391/97/160 for tieredImageNet. 
For the meta few-shot learning formation, we also follow the standard protocol: Each episode/task is formed by taking 5 random classes, and take $k=1$ or $k=5$ samples from each class for the support set $S$ in the $1$-shot or $5$-shot cases. The query set is composed of $k_q=15$ samples per class. 
We only deal with $C=5$-way classification. 
The number of meta training iterations (i.e., the number of episodes) is chosen as 600 for $1$-shot and 400 for $5$-shot problems. Similarly as~\citep{gpdkt}, the test performance is measured on 600 random test episodes/tasks averaged over $5$ random runs. 
For the cross-domain setup, we consider two problems: i) \textbf{OMNIGLOT$\to$EMNIST} (that is, trained on the OMNIGLOT dataset~\citep{omniglot} and validated/tested on the \textbf{EMNIST}~\citep{emnist}) and ii) \textbf{{\em mini}ImageNet$\to$CUB}. 
We follow the data splits, protocols, and other training details that are identical to those described in~\citep{gpdkt}.

\subsection{Results}\label{sec:results}


\subsubsection{Within-domain classification}

The results on the \textbf{CUB}, \textbf{{\em mini}ImageNet}, and \textbf{{\em tiered}ImageNet} datasets are summarized in Table~\ref{tab:cub},  Table~\ref{tab:miniimagenet}, and Table~\ref{tab:tieredimagenet}, respectively. To have fair comparison with existing approaches, we test our model on the four-layer convolutional network (known as  \textbf{Conv-4}) used in~\citep{protonet,matchingnet} and \textbf{ResNet-10} as the backbone networks for the \textbf{CUB} dataset. For the \textbf{{\em mini}ImageNet} and \textbf{{\em tiered}ImageNet}, we use the \textbf{Conv-4} and \textbf{ResNet-18}. We compare our \textbf{GPLDLA} with several state-of-the-arts, including \textbf{MAML}~\citep{maml}, \textbf{ProtoNet}~\citep{protonet}, \textbf{MatchingNet}~\citep{matchingnet}, and \textbf{RelationNet}~\citep{relationnet}. We also consider the simple feature transfer, as well as strong baselines such as \textbf{Baseline$++$}~\citep{baseline} and \textbf{SimpleShot}~\citep{simpleshot}. Among others, the (hierarchical) Bayesian approaches including \textbf{VERSA}~\citep{versa}, \textbf{LLAMA}~\citep{llama}, and \textbf{Meta-Mixture}~\citep{meta_mix}, are also compared. However, we exclude methods that use more complex backbones or more sophisticated learning schedules~\citep{ye2020fewshot,self_critique,tadam,Act2Param}, and those that require a large number of extra parameters to be trained~\citep{metaqda}.

\begin{table*}
\caption{Results on the {\em mini}ImageNet dataset. Best scores are boldfaced.}
\vspace{+0.3em}
\centering
\begin{footnotesize}
\centering
\scalebox{0.95}{
\begin{tabular}{lcccc}
\toprule
\multirow{2}{*}{Methods} & 
\multicolumn{2}{c}{Conv-4} &  \multicolumn{2}{c}{ResNet-18}
\\ \cline{2-5}
& 1-shot\Tstrut & 5-shot & 1-shot & 5-shot \\
\hline\hline
Feature Transfer\Tstrut & $39.51 \pm 0.23$ & $60.51 \pm 0.55$ & $-$ & $-$ \\
Baseline$++$~\citep{baseline}\Tstrut & $47.15 \pm 0.49$ & $66.18 \pm 0.18$ & $51.87 \pm 0.77$ & $75.68 \pm 0.63$ \\
MatchingNet~\citep{matchingnet}\Tstrut & $48.25 \pm 0.65$ & $62.71 \pm 0.44$ & $-$ & $-$ \\
ProtoNet~\citep{protonet}\Tstrut & $44.19 \pm 1.30$ & $64.07 \pm 0.65$ & $54.16 \pm 0.82$ & $73.68 \pm 0.65$ \\
MAML~\citep{maml}\Tstrut & $45.39 \pm 0.49$ & $61.58 \pm 0.53$ & $-$ & $-$ \\
RelationNet~\citep{relationnet}\Tstrut & $48.76 \pm 0.17$ & $64.20 \pm 0.28$ & $52.48 \pm 0.86$ & $69.83 \pm 0.68$ \\
ML-LSTM~\citep{mllstm}\Tstrut & $43.44 \pm 0.77$ & $60.60 \pm 0.71$ & $-$ & $-$ \\
SNAIL~\citep{snail}\Tstrut & $45.10$ & $55.20$ & $-$ & $-$ \\
VERSA~\citep{versa}\Tstrut & $48.53 \pm 1.84$ & $67.37 \pm 0.86$ & $-$ & $-$ \\
LLAMA~\citep{llama}\Tstrut & $49.40 \pm 1.83$ & $-$ & $-$ & $-$ \\
Meta-Mixture~\citep{meta_mix}\Tstrut & $49.60 \pm 1.50$ & $64.60 \pm 0.92$ & $-$ & $-$ \\
SimpleShot~\citep{simpleshot}\Tstrut & $49.69 \pm 0.19$ & $66.92 \pm 0.17$ & ${\bf 62.85 \pm 0.20}$ & ${\bf 80.02 \pm 0.14}$ \\
GPDKT$^\textrm{CosSim}$~\citep{gpdkt}\Tstrut & $48.64 \pm 0.45$ & $62.85 \pm 0.37$ & $-$ & $-$ \\
GPDKT$^\textrm{BNCosSim}$~\citep{gpdkt}\Tstrut & $49.73 \pm 0.07$ & $64.00 \pm 0.09$ & $-$ & $-$ \\
\hline
GPLDLA (Ours)\Tstrut & ${\bf 52.58 \pm 0.19}$ & ${\bf 69.59 \pm 0.16}$ & $60.05 \pm 0.20$ & $79.22 \pm 0.15$ \\
\bottomrule
\end{tabular}
}
\end{footnotesize}
\label{tab:miniimagenet}
\end{table*}

\begin{table*}
\caption{Results on the {\em tiered}ImageNet dataset. Best scores are boldfaced.}
\vspace{+0.3em}
\centering
\begin{footnotesize}
\centering
\scalebox{0.95}{
\begin{tabular}{lcccc}
\toprule
\multirow{2}{*}{Methods} & 
\multicolumn{2}{c}{Conv-4} &  \multicolumn{2}{c}{ResNet-18}
\\ \cline{2-5}
& 1-shot\Tstrut & 5-shot & 1-shot & 5-shot \\
\hline\hline
ProtoNet~\citep{protonet}\Tstrut & $53.31 \pm 0.89$ & $72.69 \pm 0.74$ & $-$ & $-$ \\
MAML~\citep{maml}\Tstrut & $51.67 \pm 1.81$ & $70.30 \pm 1.75$ & $-$ & $-$ \\
RelationNet~\citep{relationnet}\Tstrut & $54.48 \pm 0.48$ & $71.31 \pm 0.78$ & $-$ & $-$ \\
SimpleShot~\citep{simpleshot}\Tstrut & $51.02 \pm 0.20$ & $68.98 \pm 0.18$ & $69.09 \pm 0.22$ & $84.58 \pm 0.16$ \\
GPDKT$^\textrm{CosSim}$~\citep{gpdkt}\Tstrut & $51.14 \pm 0.21$ & $67.20 \pm 0.18$ & $62.65 \pm 0.23$ & $79.93 \pm 0.17$ \\
\hline
GPLDLA (Ours)\Tstrut & ${\bf 54.75 \pm 0.24}$ & ${\bf 72.93 \pm 0.26}$ & ${\bf 69.45 \pm 0.37}$ & ${\bf 85.16 \pm 0.19}$ \\
\bottomrule
\end{tabular}
}
\end{footnotesize}
\label{tab:tieredimagenet}
\end{table*}

Our approach achieves the best performance on most of the setups.
On the CUB dataset, GPLDLA attains the highest accuracies for three cases out of four. On the {\em mini}ImageNet, GPLDLA exhibits significantly higher performance than competing methods when the simpler backbone (Conv-4) is used, while being the second best and comparable to SimpleShot with the ResNet18 backbone\footnote{
SimpleShot with ResNet-18 backbone on the CUB 
scored accuracy $64.46$ (1-shot) and $81.56$ (5-shot). 
}.
And our GPLDLA outperforms GPDKT with all different kernels in most of the cases. GPLDLA also performs the best on {\em tiered}ImageNet.

\begin{table*}[t!]
\caption{Cross-domain classification performance. 
}
\centering
\begin{footnotesize}
\centering
\scalebox{0.95}{
\begin{tabular}{lcccc}
\toprule
\multirow{2}{*}{Methods} & 
\multicolumn{2}{c}{OMNIGLOT$\to$EMNIST} &  \multicolumn{2}{c}{{\em mini}ImageNet$\to$CUB}
\\ \cline{2-5}
& 1-shot\Tstrut & 5-shot & 1-shot & 5-shot \\
\hline\hline
Feature Transfer\Tstrut & $64.22 \pm 1.24$ & $86.10 \pm 0.84$ & $32.77 \pm 0.35$ & $50.34 \pm 0.27$ \\
Baseline$++$~\citep{baseline}\Tstrut & $56.84 \pm 0.91$ & $80.01 \pm 0.92$ & $39.19 \pm 0.12$ & $57.31 \pm 0.11$ \\
MatchingNet~\citep{matchingnet}\Tstrut & $75.01 \pm 2.09$ & $87.41 \pm 1.79$ & $36.98 \pm 0.06$ & $50.72 \pm 0.36$ \\
ProtoNet~\citep{protonet}\Tstrut & $72.04 \pm 0.82$ & $87.22 \pm 1.01$ & $33.27 \pm 1.09$ & $52.16 \pm 0.17$ \\
MAML~\citep{maml}\Tstrut & $72.68 \pm 1.85$ & $83.54 \pm 1.79$ & $34.01 \pm 1.25$ & $48.83 \pm 0.62$ \\
RelationNet~\citep{relationnet}\Tstrut & $75.62 \pm 1.00$ & $87.84 \pm 0.27$ & $37.13 \pm 0.20$ & $51.76 \pm 1.48$ \\
GPDKT$^\textrm{Linear}$~\citep{gpdkt}\Tstrut & $75.97 \pm 0.70$ & $89.51 \pm 0.44$ & $38.72 \pm 0.42$ & $54.20 \pm 0.37$ \\
GPDKT$^\textrm{CosSim}$~\citep{gpdkt}\Tstrut & $73.06 \pm 2.36$ & $88.10 \pm 0.78$ & $40.22 \pm 0.54$ & $55.65 \pm 0.05$ \\
GPDKT$^\textrm{BNCosSim}$~\citep{gpdkt}\Tstrut & $75.40 \pm 1.10$ & ${\bf 90.30 \pm 0.49}$ & $40.14 \pm 0.18$ & $56.40 \pm 1.34$ \\
\hline
GPLDLA (Ours)\Tstrut & ${\bf 76.65 \pm 0.29}$ & $89.71 \pm 0.14$ & ${\bf 41.92 \pm 0.27}$ & ${\bf 60.88 \pm 0.30}$ \\
\bottomrule
\end{tabular}
}
\end{footnotesize}
\label{tab:cross_domain}
\end{table*}

\subsubsection{Cross-domain classification}

Unlike within-domain classification, we test the trained model on test data from a different domain/dataset. This cross-domain experiments can judge the generalization performance of the few-shot algorithms in challenging unseen domain scenarios. The results 
are summarized in 
Table~\ref{tab:cross_domain} where we use the Conv-4 backbone for both cases. 
GPLDLA exhibits the best performance for most cases outperforming GPDKT, except for one case. Our GPLDLA also performs comparably well with recent approaches with the ResNet-18 backbone on the miniImageNet$\to$CUB task as shown in Table~\ref{tab:cross_domain_resnet18}.

\begin{table}
\caption{Cross-domain classification performance with ResNet-18 backbone on miniImageNet$\to$CUB. 
Assoc-Align = \citep{assoc_align}, 
Neg-Margin = \citep{neg_margin}, and
Cross-Domain = \citep{cross_dom}. 
}
\centering
\vspace{-0.4em}
\begin{small}
\centering
\scalebox{0.95}{
\begin{tabular}{lcc}
\toprule
Methods\Tstrut & 1-shot & 5-shot \\
\hline\hline
Assoc-Align\Tstrut & $47.25 \pm 0.76$ & ${\bf 72.37 \pm 0.89}$ \\
Neg-Margin\Tstrut & $-$ & $69.30 \pm 0.73$ \\
Cross-Domain\Tstrut & $47.47 \pm 0.75$ & $66.98 \pm 0.68$ \\
\hline
GPLDLA (Ours)\Tstrut & ${\bf 48.94 \pm 0.45}$ & $69.83 \pm 0.36$ \\
\bottomrule
\end{tabular}
}
\end{small}
\label{tab:cross_domain_resnet18}
\end{table}

\subsubsection{Calibration errors
}

Considering the practical use of the machine learning algorithms, it is important to align the model's prediction accuracy and its prediction confidence. For instance, when model's prediction is wrong, it would be problematic if the confidence of prediction is high. In this section we evaluate this alignment measure for our approach. Specifically we employ the expected calibration error (ECE)~\citep{ece} as the measure of misalignment. The ECE can be computed by the following procedure: the model's prediction confidence scores on the test cases are sorted and partitioned into $H$ bins (e.g., $H=20$), and for each bin we compute the difference between prediction accuracy (on the test examples that belong to the bin) and the confidence score of the bin. The ECE is the weighted average of these differences over the bins with the weights proportional to the numbers of bin samples. Hence the smaller the better.


Following~\citep{gpdkt}, we sample 3000 tasks from the test set on the CUB dataset, and calibrate the temperature parameter by minimizing the negative log-likelihood score, and use another 3000 tasks from the test data to evaluate the ECE loss. 
The ECE losses averaged over five random runs are summarized in Table~\ref{tab:calibration}. 
On the 1-shot case, our GPLDLA attains the lowest calibration error, while being slightly worse than ProtoNet and GPDKT on 5-shot.


\begin{table}
\centering
\caption{Expected calibration errors. 
}
\label{tab:calibration}
\vspace{-0.4em}
\begin{footnotesize}
\begin{sc}
\centering
\begin{tabular}{lcc}
\toprule
Methods\Tstrut & 1-shot & 5-shot \\
\hline\hline
Feature Transfer\Tstrut & $12.57 \pm 0.23$ & $18.43 \pm 0.16$ \\
Baseline$++$
& $4.91 \pm 0.81$ & $2.04 \pm 0.67$ \\
MatchingNet
& $3.11 \pm 0.39$ & $2.23 \pm 0.25$ \\
ProtoNet
& $1.07 \pm 0.15$ & ${\bf 0.93 \pm 0.16}$ \\
MAML
& $1.14 \pm 0.22$ & $2.47 \pm 0.07$ \\
RelationNet
& $4.13 \pm 1.72$ & $2.80 \pm 0.63$ \\
GPDKT$^\textrm{BNCosSim}$
& $2.62 \pm 0.19$ & $1.15 \pm 0.21$ \\
\hline
GPLDLA (Ours)\Tstrut & ${\bf 0.74 \pm 0.12}$ & $1.34 \pm 0.16$ \\
\bottomrule
\end{tabular}
\end{sc}
\end{footnotesize}
\vspace{-1.0em}
\end{table}

\subsection{Ablation study }\label{sec:ablation}

To verify the impact of the proposed approximation strategy of LDA plugin and prior-norm adjustment, we conduct ablation study in this section. We compare three models: i) Laplace approximation that finds the MAP solution without any approximation (neither LDA plugin nor prior-norm adjustment), ii) approximate MAP estimate by LDA plugin alone without prior-norm adjustment, and iii) both LDA plugin and prior-norm adjustment (hence our GPLDLA). 
For Laplace approximation we take 5 Newton steps to find the MAP solution, where the cascaded update operations are differentiable for meta learning of the prior parameters. For the LDA plugin alone, we need to estimate the class-conditional variance parameter $\sigma^2$, and we adopt the median distance heuristics: $\sigma$ set as the median of all pairwise feature distances (in the support set), a common practice for determining feature scales in kernel machines~\citep{med_dist}. 

Table~\ref{tab:ablation} summarizes the test performance of the three models. Overall our GPLDLA and Laplace approximation perform equally well whereas GPLDLA slightly outperforms Laplace approximation most of the time. This may be attributed to the ProtoNet-like effect of the LDA plugin estimator, which serves as an additional regularizer for few-shot learning. We also see that LDA plugin alone significantly underperforms GPLDLA, implying that prior-norm adjustment is more effective than heuristic median-distance rule in determining the class conditional feature scale. However, the differences are less pronounced for larger backbones (ResNets) in which the importance of the sophisticated features is dominant.

\begin{table*}
\vspace{+0.5em}
\caption{Ablation study comparing Laplace approximation, LDA plugin alone without prior-norm adjustment, and ours (LDA plugin $+$ prior-norm adjustment).}
\vspace{+0.3em}
\centering
\begin{footnotesize}
\centering
\scalebox{0.95}{
\begin{tabular}{lcccc}
\multicolumn{5}{c}{(a) CUB}\\
\toprule
\multirow{2}{*}{Methods} & 
\multicolumn{2}{c}{Conv-4} &  \multicolumn{2}{c}{ResNet-10}
\\ \cline{2-5}
& 1-shot\Tstrut & 5-shot & 1-shot & 5-shot \\
\hline\hline
Laplace approximation\Tstrut & $61.94 \pm 0.22$ & $78.31 \pm 0.16$ & $70.57 \pm 0.23$ & $84.62 \pm 0.13$ \\
LDA plugin alone\Tstrut & $49.27 \pm 0.23$ & $62.69 \pm 0.19$ & $52.28 \pm 0.24$ & $66.61 \pm 0.20$ \\
\hline
LDA plugin $+$ prior-norm adjustment (Ours)\Tstrut & ${\bf 63.40 \pm 0.14}$ & ${\bf 78.86 \pm 0.35}$ & ${\bf 71.30 \pm 0.16}$ & ${\bf 86.38 \pm 0.15}$ \\
\bottomrule
\end{tabular}
}
\\
\vspace{+0.5em}
\scalebox{0.95}{
\begin{tabular}{lcccc}
\multicolumn{5}{c}{(b) {\em mini}ImageNet}\\
\toprule
\multirow{2}{*}{Methods} & 
\multicolumn{2}{c}{Conv-4} &  \multicolumn{2}{c}{ResNet-18}
\\ \cline{2-5}
& 1-shot\Tstrut & 5-shot & 1-shot & 5-shot \\
\hline\hline
Laplace approximation\Tstrut & $52.47 \pm 0.19$ & $69.42 \pm 0.16$ & $59.61 \pm 0.20$ & $79.37 \pm 0.14$ \\
LDA plugin alone\Tstrut & $42.73 \pm 0.18$ & $53.29 \pm 0.16$ & $59.66 \pm 0.20$ & ${\bf 79.50 \pm 0.14}$ \\
\hline
LDA plugin $+$ prior-norm adjustment (Ours)\Tstrut & ${\bf 52.58 \pm 0.19}$ & ${\bf 69.59 \pm 0.16}$ & ${\bf 60.05 \pm 0.20}$ & $79.22 \pm 0.15$ \\
\bottomrule
\end{tabular}
}
\\
\vspace{+0.5em}
\scalebox{0.95}{
\begin{tabular}{lcccc}
\multicolumn{5}{c}{(c) {\em tiered}ImageNet}\\
\toprule
\multirow{2}{*}{Methods} & 
\multicolumn{2}{c}{Conv-4} &  \multicolumn{2}{c}{ResNet-18}
\\ \cline{2-5}
& 1-shot\Tstrut & 5-shot & 1-shot & 5-shot \\
\hline\hline
Laplace approximation\Tstrut & $53.46 \pm 0.22$ & $71.97 \pm 0.18$ & $67.93 \pm 0.22$ & $84.09 \pm 0.16$ \\
LDA plugin alone\Tstrut & $39.59 \pm 0.20$ & $50.93 \pm 0.19$ & $64.19 \pm 0.22$ & $76.16 \pm 0.22$ \\
\hline
LDA plugin $+$ prior-norm adjustment (Ours)\Tstrut & ${\bf 54.75 \pm 0.24}$ & ${\bf 72.93 \pm 0.26}$ & ${\bf 69.45 \pm 0.37}$ & ${\bf 85.16 \pm 0.19}$ \\
\bottomrule
\end{tabular}
}
\end{footnotesize}
\label{tab:ablation}
\end{table*}

\subsection{Running times }\label{sec:runtime}

Next we measure the wall clock running times for competing Bayesian meta few-shot methods. The per-episode inference time with the Conv-4 backbone on {\em tiered}ImageNet is reported in Table~\ref{tab:running_time}. It shows that GPLDLA is the fastest thanks to the efficient closed-form inference steps.  GPDKT~\citep{gpdkt} has computational overhead of solving $C$ binarized problems separately where $C$ is the number of ways. MetaQDA~\citep{metaqda} suffers from slow inference due to the cubic time (in the feature dimension) to deal with full covariance matrices and their inverses.

\begin{table}
\caption{Per-episode inference time (milliseconds) with the Conv-4 backbone on {\em tiered}ImageNet. We exclude the feature computation times and only measure the inference time, that is, time for computing $p(y|x,S)$ for $(x,y)\in Q$. 
}
\centering
\vspace{-0.4em}
\begin{small}
\centering
\scalebox{0.95}{
\begin{tabular}{lcc}
\toprule
Methods\Tstrut & 1-shot & 5-shot \\
\hline\hline
GPDKT~\citep{gpdkt}\Tstrut & $10.42 \pm 0.24$ & $12.02 \pm 0.08$ \\
MetaQDA (Full Bayesian)~\citep{metaqda}\Tstrut & $22.57 \pm 0.57$ & $25.84 \pm 1.19$ \\
MetaQDA (MAP)~\citep{metaqda}\Tstrut & $20.29 \pm 0.22$ & $22.49 \pm 0.67$ \\
Laplace approximation\Tstrut & $11.42 \pm 1.70$ & $14.08 \pm 0.63$ \\
\hline
GPLDLA (Ours)\Tstrut & ${\bf 6.70 \pm 0.03}$ & ${\bf 6.71 \pm 0.14}$ \\
\bottomrule
\end{tabular}
}
\end{small}
\label{tab:running_time}
\end{table}

\section{Conclusion}\label{sec:conclusion}

We proposed a novel GP meta learning algorithm for few-shot classification. We adopt the Laplace posterior approximation but circumvent iterative gradient steps for finding the MAP solution by our novel LDA plugin with prior-norm adjustment. This enables closed-form differentiable GP posteriors and predictive distributions, thus allowing fast meta training. We empirically verified that our approach attained considerable improvement over previous approaches in both standard benchmark datasets and cross-domain adaptation scenarios.
We would like to conclude the paper by discussing some limitations of the proposed approach. In the current study we have not carried out rigorous theoretical analysis on the impact of the cascade of approximations that we used. 
Our strategy looks reasonable, and works well on empirical data, but further theoretical study 
needs to be carried out as future work.


{
\small
\bibliography{main}

\begin{thebibliography}{42}
\providecommand{\natexlab}[1]{#1}
\providecommand{\url}[1]{\texttt{#1}}
\expandafter\ifx\csname urlstyle\endcsname\relax
  \providecommand{\doi}[1]{doi: #1}\else
  \providecommand{\doi}{doi: \begingroup \urlstyle{rm}\Url}\fi

\bibitem[Afrasiyabi et~al.(2020)Afrasiyabi, Lalonde, and
  Gagn\'{e}]{assoc_align}
Arman Afrasiyabi, Jean-François Lalonde, and Christian Gagn\'{e}.
\newblock Associative alignment for few-shot image classification.
\newblock In \emph{European Conference on Computer Vision}, 2020.

\bibitem[Antoniou and Storkey(2019)]{self_critique}
Antreas Antoniou and Amos Storkey.
\newblock Learning to learn via self-critique.
\newblock In \emph{Advances in Neural Information Processing Systems}, 2019.

\bibitem[Bengio et~al.(1992)Bengio, Bengio, Cloutier, and Gecsei]{meta_bengio}
Samy Bengio, Yoshua Bengio, Jocelyn Cloutier, and Jan Gecsei.
\newblock On the optimization of a synaptic learning rule.
\newblock In \emph{Optimality in Artificial and Biological Neural Networks},
  1992.

\bibitem[Bishop(2006)]{bishop:2006:PRML}
Christopher~M. Bishop.
\newblock \emph{Pattern Recognition and Machine Learning}.
\newblock Springer, 2006.

\bibitem[Chen et~al.(2019)Chen, Liu, Kira, Wang, and Huang]{baseline}
Wei-Yu Chen, Yen-Cheng Liu, Zsolt Kira, Yu-Chiang Wang, and Jia-Bin Huang.
\newblock A closer look at few-shot classification.
\newblock In \emph{International Conference on Learning Representations}, 2019.

\bibitem[Cohen et~al.(2016)Cohen, Afshar, Tapson, and van Schaik]{emnist}
Gregory Cohen, Saeed Afshar, Jonathan Tapson, and Andr\'{e} van Schaik.
\newblock {EMNIST: Extending MNIST to handwritten letters}.
\newblock In \emph{International Joint Conference on Neural Networks (IJCNN)},
  2016.

\bibitem[Fei-Fei et~al.(2006)Fei-Fei, Fergus, and Perona]{oneshot}
Li~Fei-Fei, R.~Fergus, and P.~Perona.
\newblock One-shot learning of object categories.
\newblock \emph{IEEE Transactions on Pattern Analysis and Machine
  Intelligence}, 28\penalty0 (4):\penalty0 594--611, 2006.
\newblock \doi{10.1109/TPAMI.2006.79}.

\bibitem[Finn et~al.(2017)Finn, Abbeel, and Levine]{maml}
Chelsea Finn, Pieter Abbeel, and Sergey Levine.
\newblock Model-agnostic meta-learning for fast adaptation of deep networks.
\newblock In \emph{International Conference on Machine Learning}, 2017.

\bibitem[Garreau et~al.(2018)Garreau, Jitkrittum, and Kanagawa]{med_dist}
Damien Garreau, Wittawat Jitkrittum, and Motonobu Kanagawa.
\newblock Large sample analysis of the median heuristic.
\newblock In \emph{arXiv preprint}, 2018.
\newblock URL \url{https://arxiv.org/abs/1707.07269}.

\bibitem[Gordon et~al.(2019)Gordon, Bronskill, Bauer, Nowozin, and
  Turner]{versa}
Jonathan Gordon, John Bronskill, Matthias Bauer, Sebastian Nowozin, and Richard
  Turner.
\newblock Meta-learning probabilistic inference for prediction.
\newblock In \emph{International Conference on Learning Representations}, 2019.
\newblock URL \url{https://openreview.net/forum?id=HkxStoC5F7}.

\bibitem[Grant et~al.(2018)Grant, Finn, Levine, Darrell, and Griffiths]{llama}
Erin Grant, Chelsea Finn, Sergey Levine, Trevor Darrell, and Thomas Griffiths.
\newblock {Recasting Gradient-Based Meta-Learning as Hierarchical Bayes}.
\newblock In \emph{International Conference on Learning Representations}, 2018.

\bibitem[Guo et~al.(2017)Guo, Pleiss, Sun, and Weinberger]{ece}
Chuan Guo, Geoff Pleiss, Yu~Sun, and Kilian~Q. Weinberger.
\newblock On calibration of modern neural networks.
\newblock In \emph{International Conference on Machine Learning}, 2017.

\bibitem[He et~al.(2015)He, Zhang, Ren, and Sun]{resnet}
Kaiming He, Xiangyu Zhang, Shaoqing Ren, and Jian Sun.
\newblock Deep residual learning for image recognition.
\newblock \emph{CoRR}, abs/1512.03385, 2015.
\newblock URL \url{http://arxiv.org/abs/1512.03385}.

\bibitem[Hospedales et~al.(2020)Hospedales, Antoniou, Micaelli, and
  Storkey]{survey_meta}
Timothy Hospedales, Antreas Antoniou, Paul Micaelli, and Amos Storkey.
\newblock {Meta-Learning in Neural Networks: A Survey}.
\newblock In \emph{arXiv preprint}, 2020.
\newblock URL \url{https://arxiv.org/abs/2004.05439}.

\bibitem[Huisman et~al.(2021)Huisman, van Rijn, and Plaat]{survey_deep_meta}
Mike Huisman, Jan~N. van Rijn, and Aske Plaat.
\newblock A survey of deep meta-learning.
\newblock In \emph{arXiv preprint}, 2021.
\newblock URL \url{https://arxiv.org/abs/2010.03522}.

\bibitem[Jerfel et~al.(2019)Jerfel, Grant, Griffiths, and Heller]{meta_mix}
Ghassen Jerfel, Erin Grant, Thomas~L. Griffiths, and Katherine Heller.
\newblock Reconciling meta-learning and continual learning with online mixtures
  of tasks.
\newblock In \emph{Advances in Neural Information Processing Systems}, 2019.

\bibitem[Lake et~al.(2013)Lake, Salakhutdinov, and Tenenbaum]{omniglot}
B.~M. Lake, R.~R. Salakhutdinov, and J.~Tenenbaum.
\newblock One-shot learning by inverting a compositional causal process, 2013.
\newblock In Advances in Neural Information Processing Systems.

\bibitem[Liu et~al.(2020)Liu, Cao, Lin, Li, Zhang, Long, and Hu]{neg_margin}
Bin Liu, Yue Cao, Yutong Lin, Qi~Li, Zheng Zhang, Mingsheng Long, and Han Hu.
\newblock Negative margin matters: Understanding margin in few-shot
  classification.
\newblock In \emph{European Conference on Computer Vision}, 2020.

\bibitem[Mishra et~al.(2018)Mishra, Rohaninejad, Chen, and Abbeel]{snail}
Nikhil Mishra, Mostafa Rohaninejad, Xi~Chen, and Pieter Abbeel.
\newblock {A Simple Neural Attentive Meta-Learner}.
\newblock In \emph{International Conference on Learning Representations}, 2018.

\bibitem[Ng and Jordan(2001)]{ng_jordan}
Andrew~Y. Ng and Michael~I. Jordan.
\newblock {On discriminative vs. generative classifiers: A comparison of
  logistic regression and naive Bayes}.
\newblock In \emph{Advances in Neural Information Processing Systems}, 2001.

\bibitem[Oreshkin et~al.(2018)Oreshkin, Rodriguez, and Lacoste]{tadam}
Boris~N. Oreshkin, Pau Rodriguez, and Alexandre Lacoste.
\newblock {TADAM: Task dependent adaptive metric for improved few-shot
  learning}.
\newblock In \emph{Advances in Neural Information Processing Systems}, 2018.

\bibitem[Patacchiola et~al.(2020)Patacchiola, Turner, Crowley, and
  Storkey]{gpdkt}
Massimiliano Patacchiola, Jack Turner, Elliot~J. Crowley, and Amos Storkey.
\newblock Bayesian meta-learning for the few-shot setting via deep kernels.
\newblock In \emph{Advances in Neural Information Processing Systems}, 2020.

\bibitem[Qiao et~al.(2018)Qiao, Liu, Shen, and Yuille]{Act2Param}
Siyuan Qiao, Chenxi Liu, Wei Shen, and Alan~L. Yuille.
\newblock Few-shot image recognition by predicting parameters from activations.
\newblock In \emph{IEEE/CVF Conference on Computer Vision and Pattern
  Recognition (CVPR)}, 2018.

\bibitem[Rasmussen and Williams(2006)]{gpml_book}
Carl~Edward Rasmussen and Christopher K.~I. Williams.
\newblock \emph{Gaussian Processes for Machine Learning}.
\newblock The MIT Press, 2006.

\bibitem[Ravi and Larochelle(2017)]{mllstm}
Sachin Ravi and Hugo Larochelle.
\newblock {Optimization as a Model for Few-Shot Learning}.
\newblock In \emph{International Conference on Learning Representations}, 2017.

\bibitem[Ren et~al.(2018)Ren, Triantafillou, Ravi, Snell, Swersky, Tenenbaum,
  Larochelle, and Zemel]{tiered}
Mengye Ren, Eleni Triantafillou, Sachin Ravi, Jake Snell, Kevin Swersky,
  Joshua~B. Tenenbaum, Hugo Larochelle, and Richard~S. Zemel.
\newblock {Meta-Learning for Semi-Supervised Few-Shot Classification}.
\newblock \emph{arXiv preprint arXiv:1803.00676}, 2018.

\bibitem[Schmidhuber(1992)]{meta_schmid}
Jürgen Schmidhuber.
\newblock Learning to control fast-weight memories: An alternative to dynamic
  recurrent networks.
\newblock \emph{Neural Computation}, 4\penalty0 (1):\penalty0 131--139, 1992.
\newblock \doi{10.1162/neco.1992.4.1.131}.

\bibitem[Snell et~al.(2017)Snell, Swersky, and Zemel]{protonet}
Jake Snell, Kevin Swersky, and Richard~S. Zemel.
\newblock Prototypical networks for few-shot learning.
\newblock \emph{CoRR}, abs/1703.05175, 2017.
\newblock URL \url{http://arxiv.org/abs/1703.05175}.

\bibitem[Steyvers et~al.(2006)Steyvers, Griffiths, and Dennis]{prob_semantic}
M.~Steyvers, T.~L. Griffiths, and S.~Dennis.
\newblock Probabilistic inference in human semantic memory.
\newblock \emph{Trends in Cognitive Sciences}, 10\penalty0 (7):\penalty0
  327--334, 2006.

\bibitem[Sung et~al.(2018)Sung, Yang, Zhang, Xiang, Torr, and
  Hospedales]{relationnet}
Flood Sung, Yongxin Yang, Li~Zhang, Tao Xiang, Philip H.~S. Torr, and
  Timothy~M. Hospedales.
\newblock {Learning to Compare: Relation Network for Few-Shot Learning}.
\newblock In \emph{IEEE/CVF Conference on Computer Vision and Pattern
  Recognition (CVPR)}, June 2018.

\bibitem[Tenenbaum et~al.(2011)Tenenbaum, Kemp, Griffiths, and
  Goodman]{grow_mind}
Joshua~B. Tenenbaum, Charles Kemp, Thomas~L. Griffiths, and Noah~D. Goodman.
\newblock How to grow a mind: Statistics, structure, and abstraction.
\newblock \emph{Science}, 331\penalty0 (6022):\penalty0 1279--1285, 2011.

\bibitem[Thrun and Pratt(1998)]{learn2learn}
Sebastian Thrun and Lorien Pratt.
\newblock \emph{Learning to learn}.
\newblock Kluwer Academic Publishers, 1998.

\bibitem[Tseng et~al.(2020)Tseng, Lee, Huang, and Yang]{cross_dom}
Hung-Yu Tseng, Hsin-Ying Lee, Jia-Bin Huang, and Ming-Hsuan Yang.
\newblock Cross-domain few-shot classification via learned feature-wise
  transformation.
\newblock In \emph{International Conference on Learning Representations}, 2020.

\bibitem[Vinyals et~al.(2016)Vinyals, Blundell, Lillicrap, Kavukcuoglu, and
  Wierstra]{matchingnet}
Oriol Vinyals, Charles Blundell, Timothy Lillicrap, Koray Kavukcuoglu, and Daan
  Wierstra.
\newblock Matching networks for one shot learning.
\newblock In \emph{Advances in Neural Information Processing Systems}, 2016.

\bibitem[Wang et~al.(2019)Wang, Chao, Weinberger, and van~der
  Maaten]{simpleshot}
Yan Wang, Wei-Lun Chao, Kilian~Q. Weinberger, and Laurens van~der Maaten.
\newblock {SimpleShot: Revisiting Nearest-Neighbor Classification for Few-Shot
  Learning}.
\newblock \emph{arXiv preprint arXiv:1911.04623}, 2019.

\bibitem[Wang et~al.(2020)Wang, Yao, Kwok, and Ni]{survey_few_shot}
Yaqing Wang, Quanming Yao, James Kwok, and Lionel~M. Ni.
\newblock {Generalizing from a Few Examples: A Survey on Few-Shot Learning}.
\newblock In \emph{arXiv preprint}, 2020.
\newblock URL \url{https://arxiv.org/abs/1904.05046}.

\bibitem[Welinder et~al.(2010)Welinder, Branson, Mita, Wah, Schroff, Belongie,
  and Perona]{cub}
P.~Welinder, S.~Branson, T.~Mita, C.~Wah, F.~Schroff, S.~Belongie, and
  P.~Perona.
\newblock {Caltech-UCSD Birds 200}.
\newblock Technical Report CNS-TR-2010-001, California Institute of Technology,
  2010.

\bibitem[Wilson et~al.(2016)Wilson, Hu, Salakhutdinov, and Xing]{dkl16}
Andrew~Gordon Wilson, Zhiting Hu, Ruslan Salakhutdinov, and Eric~P. Xing.
\newblock Deep kernel learning, 2016.
\newblock AI and Statistics (AISTATS).

\bibitem[Xu et~al.(2020)Xu, Ton, Kim, Kosiorek, and Teh]{metafun}
Jin Xu, Jean-Francois Ton, Hyunjik Kim, Adam~R. Kosiorek, and Yee~Whye Teh.
\newblock {MetaFun: Meta-Learning with Iterative Functional Updates}.
\newblock In \emph{International Conference on Machine Learning}, 2020.

\bibitem[Ye et~al.(2020)Ye, Hu, Zhan, and Sha]{ye2020fewshot}
Han-Jia Ye, Hexiang Hu, De-Chuan Zhan, and Fei Sha.
\newblock Few-shot learning via embedding adaptation with set-to-set functions.
\newblock In \emph{IEEE/CVF Conference on Computer Vision and Pattern
  Recognition (CVPR)}, pages 8808--8817, 2020.

\bibitem[Yoon et~al.(2018)Yoon, Kim, Dia, Kim, Bengio, and Ahn]{bmaml}
Jaesik Yoon, Taesup Kim, Ousmane Dia, Sungwoong Kim, Yoshua Bengio, and Sungjin
  Ahn.
\newblock {Bayesian Model-Agnostic Meta-Learning}.
\newblock In \emph{Advances in Neural Information Processing Systems}, 2018.

\bibitem[Zhang et~al.(2021)Zhang, Meng, Gouk, and Hospedales]{metaqda}
Xueting Zhang, Debin Meng, Henry Gouk, and Timothy Hospedales.
\newblock {Shallow Bayesian Meta Learning for Real-World Few-Shot Recognition}.
\newblock \emph{arXiv preprint arXiv:2101.02833}, 2021.

\end{thebibliography}
\bibliographystyle{plainnat}
}


\appendix




\section{Detailed Derivations }\label{sec:derivations}

In this section we derive the Laplace approximation of the GP posterior $p(W,B|S)$, which has the results (11--13) in the main paper.

\subsection{Laplace approximation of the GP posterior}

In the Laplace approximation, we approximate the posterior by a Gaussian with mean equal to the mode of the posterior (i.e., MAP) and covariance equal to the negative inverse Hessian of the log-posterior function. That is, $p(W,B|S) \approx \mathcal{N}([W,B]; [W^*,B^*], -H_*^{-1})$, where $W^*,B^* = \arg\max_{W,B} \log p(W,B|S)$ and $H_* = \nabla_{W,B}^2 \log p(W,B|S) \vert_{W=W^*,B=B^*}$.  By further imposing diagonal Hessian approximation, we have:
\begin{align}
& p(W,B|S) \approx \prod_{j=1}^C \mathcal{N}(w_j; w^*_j, V^*_j) \mathcal{N}(b_j; b^*_j, v^*_j), \\
& V^*_j = \textrm{Diag}\big( -\nabla_{w_j}^2 \log p (W^*,B^*|S) \big)^{-1}, \ \ \ \ 
v^*_j = \big( -\nabla_{b_j}^2 \log p (W^*,B^*|S) \big)^{-1},
\end{align}
where $\textrm{Diag}(\cdot)$ is the matrix diagonalization operator. 
Now we derive the Hessians. Recalling that the log-posterior $\log p(W,B|S)$ 
is written, up to constant, as (c.f.~(9) in the main paper):
\begin{align}
\sum_{(x,y)\in S} \bigg( w_y^\top\phi(x) + b_y - \log \sum_{j=1}^C e^{w_j^\top\phi(x)+b_j} \bigg) 
- \sum_{j=1}^C \bigg( \frac{||w_j||^2}{2\beta^2} + \frac{b_j^2}{2\beta_b^2} \bigg),
\label{eq:post_wb_given_S}
\end{align}
we have:
\begin{align}
& -\nabla_{w_j}^2 \log p (W,B|S) \ = \  \frac{I}{\beta^2} + \sum_{(x,y)\in S} \nabla_{w_j}^2 \log \sum_{j=1}^C e^{w_j^\top\phi(x)+b_j} \label{eq:Hw_1} \\
& \ \ \ \ \ \ \ \ \ \ \ = \ \frac{I}{\beta^2} + \sum_{(x,y)\in S} \phi(x) \phi(x)^\top \Bigg( 
\frac{e^{w_j^\top\phi(x)+b_j}}{\sum_{j'} e^{w_{j'}^\top\phi(x)+b_{j'}}} - \bigg( \frac{e^{w_j^\top\phi(x)+b_j}}{\sum_{j'} e^{w_{j'}^\top\phi(x)+b_{j'}}} \bigg)^2 
\Bigg) \label{eq:Hw_2} \\
& \ \ \ \ \ \ \ \ \ \ \ = \ \frac{I}{\beta^2} + \sum_{(x,y)\in S} \phi(x) \phi(x)^\top \Big( 
p(y=j|F(x)) - p(y=j|F(x))^2. 
\Big) \label{eq:Hw_3}
\end{align}
Now taking the $\textrm{Diag}(\cdot)$ and inverse operators, we obtain:
\begin{align}
V^*_j = \textrm{Diag}\bigg( \frac{1}{\beta^2} + \sum_{(x,y)\in S} \big( p(y=j|F^*(x)) - p(y=j|F^*(x))^2 \big) \ \phi(x)^2
\bigg)^{-1}, 
\end{align}
where $F^*(x) = \{f^*_j(x)\}_j$ with $f^*_j(x) = {w^*_j}^\top \phi(x) + b^*_j$, and all operations are elementwise. In the similar fashion, 
\begin{align}
v^*_j = \bigg( \frac{1}{\beta_b^2} + \sum_{(x,y)\in S} \big( p(y=j|F^*(x)) - p(y=j|F^*(x))^2 \big)\bigg)^{-1}. 
\end{align}

\section{Training Details}\label{sec:expmt_setup}


During the meta training we jointly optimize the GP prior parameters $\theta,\beta,\beta_b$, where $\theta$ (the parameters of the feature extractor) is initialized randomly and $\beta,\beta_b$ are initialized as $1.0$. We use the Adam optimizer with learning rate $0.002$ for $\theta$ and $0.005$ for $\beta,\beta_b$ throughout the experiments. We use the learning rate schedule with the decay step size $5$ and multiplicative factor $0.5$. 
The number of Monte Carlo samples for the GP predictive distribution is $10$. 

The Conv-4 backbone has four repeated compositions of $conv(c_{in}, c_{out})$, batch normalization, ReLU nonlinearity, and $pool(k)$, followed by the final layer flattening, where $conv(c_{in}, c_{out})$ is the convolutional layer with the numbers of input and output channels $c_{in}$ and $c_{out}$, respectively, and $pool(k)$ is the max-pooling with the filter size $k$. We have $k=2$ and $c_{in}=c_{out}=64$ except for the first layer $c_{in}=3$.

\end{document}